\documentclass[letterpaper]{article}
\usepackage{aaai20}
\usepackage{times}
\usepackage{helvet}
\usepackage{courier}
\usepackage[hyphens]{url} 
\usepackage{graphicx}
\usepackage{multirow}
\usepackage{booktabs}
\usepackage{svg}
\usepackage{amssymb}
\usepackage{eqnarray,amsmath}

\title{Importance-Aware Learning for Neural Headline Editing}

\author{
    Qingyang Wu \textsuperscript{\rm 1},
    Lei Li \textsuperscript{\rm 2},
    Hao Zhou \textsuperscript{\rm 2},
    Ying Zeng \textsuperscript{\rm 2},
    Zhou Yu \textsuperscript{\rm 1}, \\
    \textsuperscript{\rm 1}University of California, Davis,
    \textsuperscript{\rm 2}ByteDance, \\
    \{wilwu, joyu\}@ucdavis.edu, \{lileilab,zhouhao.nlp,zengying.ss\}@bytedance.com \\
}

\newcommand{\hl}{}

\begin{document}

    \maketitle
    \begin{abstract}
        Many social media news writers are not professionally trained. 
        Therefore, social media platforms have to hire professional editors to adjust amateur headlines to attract more readers. 
        We propose to automate this headline editing process through neural network models to provide more immediate writing support for these social media news writers.
        To train such a neural headline editing model, we collected a dataset which contains articles with original headlines and professionally edited headlines.
        However, it is expensive to collect a large number of professionally edited headlines.
        To solve this low-resource problem, we design an encoder-decoder model which leverages large scale pre-trained language models.
        We further improve the pre-trained model's quality by introducing a headline generation task as an intermediate task before the headline editing task.
        Also, we propose Self Importance-Aware (SIA) loss to address the different levels of editing in the dataset by down-weighting the importance of easily classified tokens and sentences.
        With the help of Pre-training, Adaptation, and SIA, the model learns to generate headlines in the professional editor's style.
        Experimental results show that our method significantly improves the quality of headline editing comparing against previous methods.
    \end{abstract}

    \section{Introduction}


    For most social media, headlines are often the first and only impression to attract readers, and people determine whether to read the article or not based on an instant scan of it.
    In other words, a good headline will result in many more views. With more views, the writer also receives more commission of advertising revenue from the social media platform.
    To catch the eyes of readers, the headlines have to be as intriguing as possible. Sometimes, a headline has no choice but to be a ``click-bait."
    Large media operations usually hire professional editors to edit headlines.
    These professional editors fix different types of problems that may happen in a headline, including grammatical errors, vague topics, lousy sentence structure, or just simply unattractive  headlines.
    After editing, those well-polished headlines typically receive more attention from social network readers.



    \begin{figure}[t]
        \raggedright
        \includegraphics[width=0.45\textwidth]{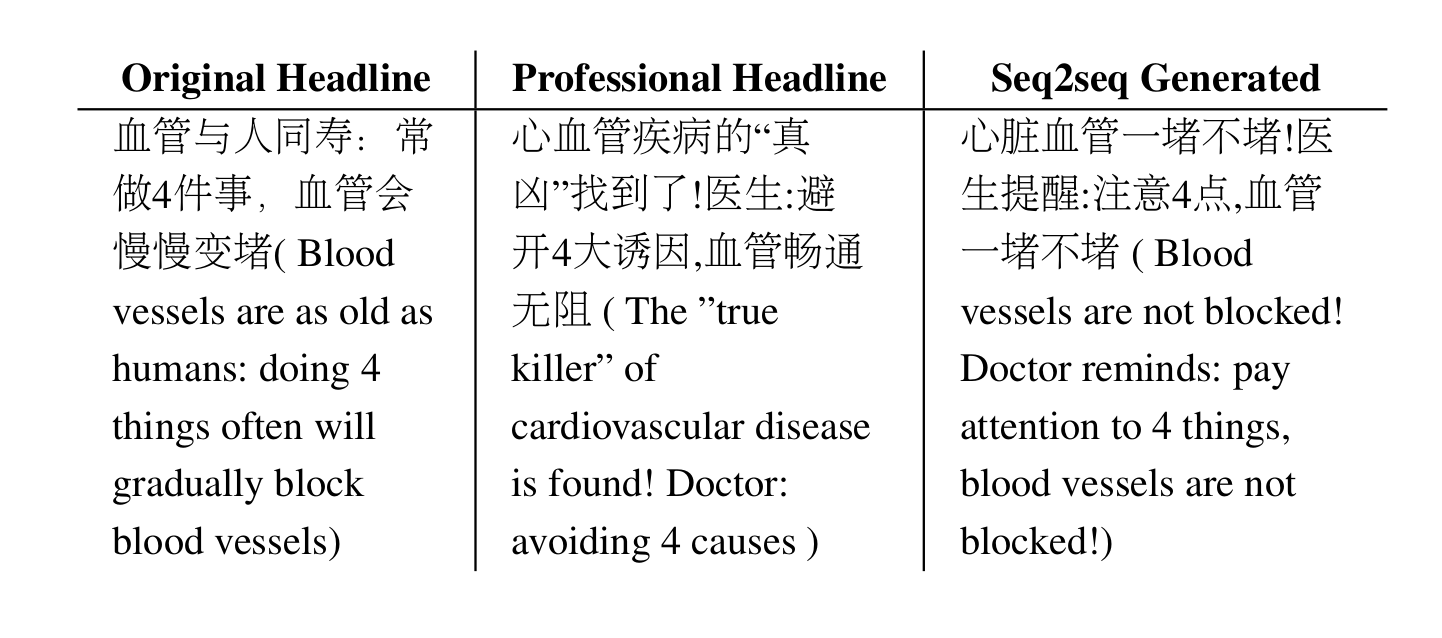}
        \caption{An example from a professional editor. The editor rewrites the original headline. A standard sequence-to-sequence model cannot produce a satisfactory result.}
        \label{tab:edit}
    \end{figure}

    \begin{figure*}[h]
        \centering
        \includegraphics[width=1.0\textwidth]{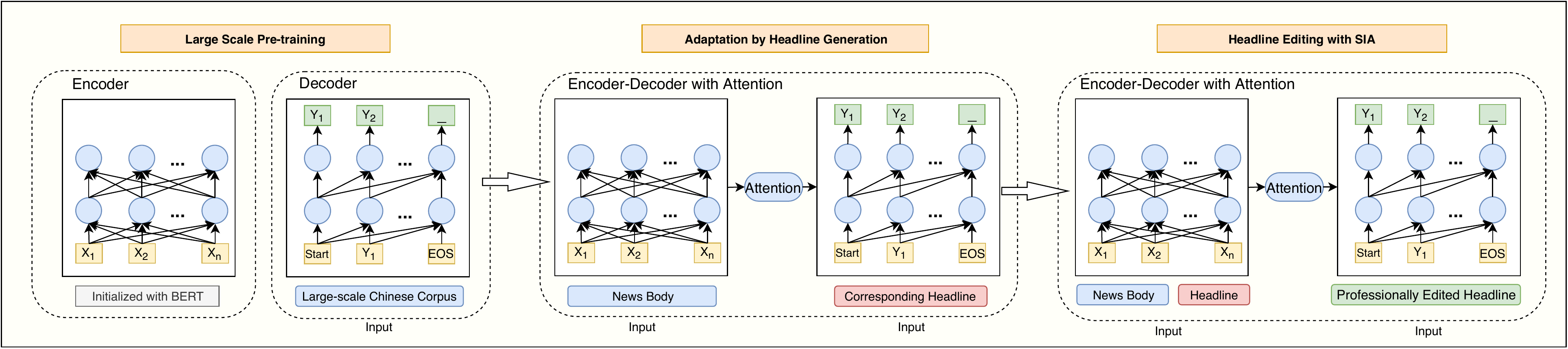}
        \caption{Overview of the training paradigm. We first pre-train the decoder as a language model using a large-scale corpus.
            Then we adapt the model using the headline generation task to train the encoder and decoder jointly before fine-tuning.
            Fine-tuning on the professional headline editing (PHED) dataset applies Self Importance-Aware loss.}
        \label{fig:model}
    \end{figure*}

    However, individual writers cannot afford to hire professional editors. Even if their articles contain exciting content, if their headlines are not attractive, people will not read them.
    For those writers, an automatic headline editing tool can be handy.
    But the task of headline editing is different from existing headline generation approaches.
    Headline generation only requires an article body to generate a headline, whereas headline editing requires an original headline as an additional input.
    Because no existing dataset contains both original headlines and professionally edited headlines,
    we construct a dataset for the headline editing task.
    We did not crowd-source the data collection since the headlines have to be written by professional editors to ensure quality.
    Therefore, we collaborate with several professional editors to create the headline editing dataset.
    Due to the limited number of professional editors, we only manage to collect 20,994 news articles with professionally edited headlines.
    We will elaborate on details of this dataset in the Dataset section.





    However, such a carefully-collected dataset does not have enough training data to build a competent neural headline editing model.
    Low-resource is a common challenge in training deep neural networks.
    In most cases, we can never have enough training data.
    Many existing studies on headline generation \cite{DBLP:conf/naacl/MuraoKKYMHT19,DBLP:journals/jcst/AyanaSLTZLS17} are data-hungry and cannot generate satisfactory results in real applications.
    Low resource also leads to text degeneration, such as repetition.
    Table~\ref{tab:edit} shows an example in the dataset and the headline generated from the standard sequence-to-sequence (Seq2seq) model.
    The generated headline here has very poor quality and contains a repetition.
    Also, notice that neural editor models \cite{DBLP:journals/tacl/GuuHOL18} that only need a small amount of data do not apply to our task because they mostly deal with situations in which two or three words are modified.
    But in our task, the professional editors often rewrite the original headline which causes a large number of edits.
    This situation forces us to favor more general generative models.

    To tackle the above problems, we leverage the large-scale pre-trained language models.
    However, since currently there is no public Generative Pre-Training (GPT) model \cite{radford2019language} for Chinese, we trained a GPT model with a large scale Chinese corpus, namely the Chinese-GPT, and we will release it to facilitate Chinese language model research.
    For the headline editing task, we build an encoder-decoder model in which the encoder is a bidirectional Transformer based on the Chinese BERT, and the decoder is the Chinese-GPT.
    In addition, we use a headline generation task as an intermediate auxiliary task before fine-tuning the model for the headline editing task.
    We also find that the imbalanced number of edits in the PHED dataset causes the model to learn easy patterns, which produces many repetitions.
    Therefore, we propose a Self Importance-Aware (SIA) loss to assign different importance for each sample, in order to prioritize the learning of harder examples.
    Experimental results show that our methods can improve both automatic and human evaluation metrics.

    Our contributions can be summarized as follows:

    \begin{itemize}
        \item We first construct a professional headline editing (PHED) dataset consisting of original headlines written by individual writers and edited headlines by professional editors.

        \item To tackle the low-resource issue, we train the model with three stages.  We first pre-train a Transformer encoder-decoder model
              and then adapt the model to a headline generation task before fine-tuning on the PHED dataset.

        \item We propose a Self Importance-Aware loss function to assign higher importance for hard examples.
    \end{itemize}

    \section{Related Work}

    \subsection{Headline Editing}
    There are many existing studies on neural headline generation \cite{DBLP:conf/acl/SunZZJ15,DBLP:journals/jcst/AyanaSLTZLS17}.
    However, their approaches and datasets only focus on generating a headline given the article body.
    In the real world, most written articles already have a corresponding headline, which makes those models less useful.
    More recent work \cite{DBLP:conf/naacl/MuraoKKYMHT19} proposes a new headline editing task, in which the goal is to create a shorter
    headline from the original headline.
    However, many writers want a more powerful tool than just making the title shorter.
    After working with professional editors, we propose a new headline editing task to address this problem.
    In the new task, the machine is responsible for editing the original headline and generating a new headline in professional style.
    It helps less-skilled writers to improve their headlines.
    Most of the articles are collected from social media, which includes different domains.
    Consequently, this task is more open-ended than the previous tasks, and therefore more challenging.

    \subsection{Large Scale Pre-trained Models}

    The recent success of BERT \cite{devlin-etal-2019-bert} and GPT2 \cite{radford2019language} have shown that such
    pre-trained language models on large-scale corpora can learn rich prior language knowledge.
    There is also a study \cite{DBLP:journals/corr/abs-1905-12616} on generating fake news that can fool humans based on large-scale pre-training.
    Therefore, to build a model that can generate human-comparable headlines, we adopt the idea of pre-training to train our own Chinese-GPT model.
    Compared with traditional methods, large scale pre-trained language models can have larger vocabularies and generalize better to unseen data.
    We further extend the pre-training by using headline generation as an intermediate auxiliary task to adapt the language model to the headline editing task.

    \subsection{Data Re-weighting}
    A common problem in natural language generation is the lack of diversity, especially when using beam-search \cite{DBLP:journals/corr/abs-1904-09751}.
    The lack of diversity causes the generated sentences to be generic and dull, often shown as repetitive patterns on both the token level and sentence level.
    This phenomenon is also related to neural text degeneration in the work of \citeauthor{DBLP:journals/corr/abs-1904-09751}.
    For example, in dialogue generation, there can be extensive generations of vague responses such as ``I don't know", ``OK", ``Sure", etc, which cause high rate of sentence-level repetitive patterns.
    The rate of such repetitions is much higher than what normally happens in human utterances.
    Existing works, such as \citeauthor{du-black-2019-boosting} (2019), use boosting methods to re-weight each data point
    by designing a rule-based discriminator to down-weight frequent repetitive responses.
    However, maintaining a list of the most frequent responses and re-calculating the similarity every time during the training consumes huge computational resources.
    Inspired by Focal Loss \cite{DBLP:conf/iccv/LinGGHD17}, we choose to let the model itself re-weight each data point dynamically during the training by evaluating the token-level and sentence-level confidence.
    In this way, we avoid maintaining a separate list of trained examples, which is
    more computationally efficient.

    \section{Methods}

    We build an encoder-decoder model in the Transformer architecture \cite{DBLP:journals/corr/VaswaniSPUJGKP17}.
    It takes the news body and the original headline as inputs to generate the professional headline.
    To solve low-source problem and repetitions, we leverage pre-training and adaptation; moreover, we propose a Self Importance-Aware loss objective.
    We use the acronym \textbf{PAS}, which stands for Pre-training, Adaptation, and SIA, to name our final model.
    Figure~\ref{fig:model} describes the overall pipeline of PAS.

    \subsection{Pre-training and Adaptation}

    The limited amount of data in the headline editing dataset hurts the model's generalizability to produce high-quality headlines on the test set.
    Therefore, we apply pre-training preceding to the headline editing task.
    It involves two stages: pre-training and adaptation.
    Each stage reduces the distance between the model's output distribution and the professionally edited headline distribution,
    which eases the difficulty in learning to generate professional headlines.
    We build a new encoder-decoder model and pre-train the encoder and decoder separately, treating them as general language models.
    We train the encoder-decoder model jointly on a headline generation dataset as adaptation before fine-tuning on the professionally edited headline dataset (PHED).

    \textbf{Pre-training}:
    The purpose of pre-training is to learn language priors that helps text generation.
    Since Chinese BERT is an available large-scale pre-trained language model, we leverage it as the pre-trained encoder here.
    However, BERT is not designed for language generation task and lacks the training on multi-domain news corpora.
    Therefore, we need to train our own autoregressive language model via generative pre-training (GPT) \cite{radford2019language}.
    We let the decoder have the same structure as BERT but apply lower-triangular mask for autoregressive text generation.
    Also to save the huge amount of time required for pre-training, we initialize the decoder with BERT's weights.
    Then, we pre-train the decoder with a MLE objective on a large-scale multi-domain Chinese corpus we collected. Details of the corpus are in the  Experiments section.
    The resulting model consists of a bidirectional Transformer as the encoder, a unidirectional Transformer as the decoder, and an attention mechanism to connect them.
    We have released the pre-trained Chinese-GPT model. \footnote{\url{http://https://github.com/qywu/Chinese-GPT}}.

    \textbf{Adaptation}:
    After pre-training, we want the model to learn the ability to summarize so that it can generate a headline given the news body.
    We train the encoder-decoder jointly with MLE using the collected headline generation dataset.
    One benefit of the joint training is that it helps to fuse the separately pre-trained encoder and decoder.
    Also, the headline generation task adapts the model to better generate headlines which lowers the difficulty of learning headline editing.

    \subsection{SIA: Self Importance-Aware Loss}

    Maximum likelihood estimation (MLE) is known to have a neural text degeneration problem \cite{DBLP:journals/corr/abs-1904-09751}.
    The generated sentences are generic and consist of repetitive patterns.
    The situation even becomes worse when using beam search decoding, as it produces many repetitions both inside a sentence and comparing to other sentences. \cite{DBLP:journals/corr/abs-1904-09751}

    We suspect that this degeneration occurs because MLE treats every data point equally, which causes it to favor easy examples more than hard ones.
    Therefore, we attribute such repetitions to the incorrect importance assignment of each data point and the internal difficulty-level imbalance across examples.
    The repetitions become worse especially in our professionally edited headline dataset, which has different levels of editing.
    To solve the problem, ideally headlines that are more specific should weight more than the ones that are general.
    However, in practice, it is difficult to decide which headline is more important than another because of the ambiguous definition of importance.

    Therefore, we propose Self Importance-Aware (SIA) loss, which lets the model itself decide how to assign the importance weight to each example.
    We use model's confidence score in predicting the ground truth to define the importance.
    A larger confidence score should result in lower importance for the target sequence.
    It is similar to the setting of perplexity, but here the range of confidence is bounded between zero and one, which enables the model to automatically adjust.

    We further notice that confidence comes from two dimensions: token-level and sentence-level:

    \begin{eqnarray}
        \text{token-level confidence} &=& p(y_t | y_{< t}) \\
        \text{sentence-level confidence} &=& \prod_{t=1}^T \, p(y_t | y_{< t})
    \end{eqnarray}
    where $y_t$ means the ground truth token at time-step $t$. $T$ represents the sequence length.
    Token-level confidence reflects how well the model predicts the next token, while sentence-level confidence represents the confidence in predicting the entire sequence.

    Next, we use confidence to decide the importance of each data point (in token-level and sentence-level), where a high confidence should result in a lower importance.
    This helps the model to learn hard examples better rather than over-fitting on easy ones.
    We can define two modulating factors: $w_t$ for the token-level importance, and $w_s$ for the sentence-level importance:

    \begin{eqnarray}
        w_t &=& (1 - p(y_t | y_{< t}))^\alpha \\
        w_s &=& (1 - \prod_{t=1}^T \, p(y_t | y_{< t}))^\beta
    \end{eqnarray}
    with $\alpha \ge 0$ and $\beta \ge 0$, where $\alpha$ and $\beta$ are two tunable hyper-parameters to decide the degree of down-weighting the importance.
    \hl{This formulation gives better numerical stability than other possible ones ($1/p$ etc.) which serve the same function.}

    Next we can combine $w_t$ and $w_s$ with the widely used MLE objective.
    This allows us to adjust the loss according to the token-level and sentence-level importance.
    Finally, we put everything together to form the SIA objective:

    \begin{equation}
        \mathcal{L_{\textrm{SIA}}} = - w_s \sum_{t=1}^T w_t \: log \, p(y_t | y_{<t})
    \end{equation}
    where $w_t$ and $w_s$ calculate the importance in token-level and sentence-level separately to down-weight easily predicted examples.
    This loss function can be used similarly to MLE during training, without any additional changes to other structure.
    It is also computationally more efficient compared to data re-weighting or boosting methods because we do not need to maintain a list of previously generated sentences for calculating the repetition.



    \section{Professional Headline Editing Dataset (PHED)}


    To generate high-quality news headlines leveraging original headlines, we propose the Professional Headline Editing Dataset (PHED). As online news headlines have varied quality, we recruited professional editors to edit original headlines to appear more attractive to readers.
    These editors are experienced social media editors that not only have good writing skills but also understand what type of headlines lead to higher numbers of views.
    Ten editors worked on 20,994 articles from six different domains including Sports (8,837), Health (7,298), Finance (1,029), Parenting (1,283), Technology (1,419), and Electronics (1,128).
    We did not impose any constraints on editing; therefore, editors could decide to rewrite the headline for better quality.
    This freedom causes different levels of editing across the data, in which 25.8\% of \hl{the total headlines} are entirely re-written and 27.3\% of \hl{the total headlines} are half edited.
    Such diverse editing makes it harder for the automatic model to learn the style of the professional headlines, suggesting rule-based methods are not applicable. Therefore, we propose an end-to-end approach to perform headline editing.

    \section{Experiments}
    Before directly building a generative model, to test the feasibility of the headline editing task, we wanted to find out if automatic models can distinguish between original headlines and professionally edited headlines.
    We fine-tuned Google Chinese BERT to perform a binary classification on original and edited headlines.
    The results show that BERT has achieved 92.56\% accuracy on the test set in finding the edited headline.
    The high accuracy suggests that the professionally edited headlines do share a ``style" that the generation model might be able to learn.
    Then we applied the proposed training paradigm for the Transformer encoder-decoder, in which we first pre-trained on the collected large-scale Chinese corpus, then adapted the model using the headline generation task, and finally fine-tuned the adapted model on the PHED dataset.
    In this section, we first introduce the collected Chinese corpus, then explain the experimental setup, and finally describe the automatic evaluation metrics to evaluate our model.

    \begin{table*}[h]
        \begin{center}
            \resizebox{0.9\textwidth}{!}{
                \begin{tabular}{l|c|c|c|c|c|c}
                    \toprule 
                    \textbf{Models}        & \textbf{Perplexity} $\downarrow$ & \textbf{BLEU-4} $\uparrow$ & \textbf{ROUGE-L}$\uparrow$ & \textbf{Token-REP-4(\%)} $\downarrow$ & \textbf{Sent-REP-4(\%)} $\downarrow$ & \textbf{Unique 4-grams} $\uparrow$ \\
                    \midrule 
                    \hl{Seq2seq}                & 21.89                            & 15.69                      & 33.46                      & 0.8206                                & 21.33                                & 12.98k                             \\
                    \hl{Seq2seq + Adapation}  & 24.45 & 13.11 & 31.14 & 2.686 & 22.30 & 12.01k \\
                    \midrule 
                    PAS                    & 7.616                            & \textbf{21.46}             & \textbf{39.47}             & \textbf{0.3722}                       & \textbf{8.882}                       & \textbf{18.20k}                    \\
                    \hl{PAS  - SIA}          & \textbf{7.588}                   & 21.13                      & 39.27                      & 0.4004                                & 9.673                                & 18.10k                             \\
                    \hl{PAS - Adaption}      & 9.273                            & 19.23                      & 37.05                      & 0.5398                                & 14.61                                & 15.58k                             \\
                    \hl{PAS  - SIA - Adaption} & 9.272                            & 18.95                      & 36.82                      & 0.5165                                & 14.91                                & 15.48k                             \\
                    \midrule 
                    \hl{Human(Expert)}                  & N/A                              & \hl{100.0}                        & \hl{100.0}                        & 0.0293                                & 7.792                                & 21.97k                             \\
                    \hl{Human(Original)} & N/A & 16.99 & 34.97 & 0.0745 & 0.8744 & 22.44k \\
                    \bottomrule 
                \end{tabular}

            }
        \end{center}
        \caption{Performance results of all models.
            Across models, we compare fine-tuning at different stages with different loss functions.
            MLE stands for fine-tuning with maximum-likelihood estimation, while SIA stands for fine-tuning with Self Importance-Aware loss. Please refer to Automatic Evaluation Metrics section for each metrics explanation.
            * SIA is set with $\alpha=0.2$ and $\beta=40$ here.}
        \label{tab:automatic-evaluation}
    \end{table*}

    \subsection{Large Scale Chinese Corpus for NLP}

    Since there is no open-source large-scale pre-trained Chinese language model available, we have to train our own Chinese language model with Transformer.
    Training such a language model requires a large-scale open-domain corpus. 
    We collect our corpus from \textit{Large Scale Chinese Corpus for NLP}
    \footnote{\url{http://github.com/brightmart/nlp_chinese_corpus}}.
    In detail, we use the following Chinese text corpus:

    \begin{itemize}
        \item \textbf{Chinese Wikipedia} (wiki2019zh) is the data to train BERT
              but we use it to train our general language model for text generation. (1.6GB)

        \item \textbf{News} (news2016zh) contains 2.5 million news articles from 63,000 sources.
              It has headlines, metadata, and body. (9.0GB)

        \item \textbf{Baike QA} (baike2018qa) is a high quality wiki question answering dataset with 493 different domains.
              It has approximately 1.5 million QA pairs. (1.1GB)

        \item \textbf{Community QA} (webtext2019zh) is similar to the WebText corpus described in \cite{radford2019language}.
              It filters text with at least 3 up-votes from 14 million comments to improve the quality.
              The final dataset contains 4.1 million comments and 28 thousand topics. (3.7GB)
    \end{itemize}
    In total, we have nearly 15.4 GB of data for language model pre-training.
    The purpose of pre-training is to learn general language priors before fine-tuning in down-stream text generation tasks.
    Moreover, we collected 332,216 news articles with news body and headlines \hl{from a news platform Toutiao} for the headline generation task.
    The process is simple \hl{because} most articles come with a corresponding headline.
    Later, we use this headline generation data for adaptation.

    \subsection{Experimental Setup}

    The encoder and decoder structures are similar to BERT, which is a 12-layer, 768 hidden size transformer.
    The difference is that we re-implement a cached version of Transformer with state reuse, mentioned in \citeauthor{DBLP:journals/corr/abs-1901-02860} (2019).
    To be consistent with BERT, we apply the exact same Chinese WordPiece tokenization in which the vocabulary has 21,128 tokens (word pieces).
    We clean all the data by replacing some punctuation and removing noises such as URLs and phone numbers.
    \hl{We add a delimiter to separate the news body and the original headline.}

    For  pre-training and adaptation, we use the standard maximum-likelihood objective.
    We stop the pre-training when the validation perplexity starts to increase.
    For fine-tuning on PHED, we switch to the proposed SIA objective.
    We conduct hyper-parameters search for finding the best $\alpha$ and $\beta$ for SIA.
    In order to reduce the variance due to randomness and ensure reproducibility, we fix the random seed across all experiments, which means that during training the batches will be fed in the same order.

    The baseline models are the sequence-to-sequence (Seq2seq) model with attention and our Transformer encoder-decoder model with only language model pre-training.
    Seq2seq uses pre-trained Chinese character embedding plus LSTM as encoder and decoder.
    We tried to set the baseline using the same tokenizer as BERT.
    However, without pre-training, it is hard for the Seq2seq to achieve good results with a large vocabulary.
    Instead, we choose 4,260 characters with frequency larger than three as the vocabulary.
    Also, we use character-level tokenization over word-level tokenization because Chinese character-based models are tested to yield better results \cite{li-etal-2019-word-segmentation}
    and also because Chinese BERT is character-based.

    During inference, we apply beam search decoding with beam size 10 for all models.
    We add the length normalization technique \cite{DBLP:journals/corr/WuSCLNMKCGMKSJL16} to \hl{deal with the fact of different lengths in the headlines as beam search would favor shorter generations}.
    The temperature is set to be $1.0$ as it yields the best result.
    Because Seq2seq uses a different tokenization method, we translate all generated tokens back into text space, so as to ensure a fair comparison.

    \subsection{Automatic Evaluation Metrics}

    We apply several automatic evaluation metrics to assess the performance of adaptation and self importance-aware loss (SIA).
    We adopt traditional automatic evaluation metrics such as \textbf{Perplexity}, \textbf{BLEU-4} \cite{DBLP:conf/acl/PapineniRWZ02} and \textbf{ROUGE-L} \cite{lin-2004-rouge}.
    Perplexity measures how well the model predicts the ground truth.
    BLEU-4 evaluates the 4-gram overlap between the generated output and the reference, and therefore gives an approximate quality evaluation of the generations.
    ROUGE-L computes the rate of the length of the largest common sub-sequence; and we include ROUGE-L because it often appears in the summarization tasks.

    More than the traditional evaluation metrics, we apply three special metrics to measure repetition and diversity for the generated outputs:

    \textbf{Token-level Repetition (Token-REP-4):}
    We define token-level repetition as the repetition \textbf{inside a sentence}.
    This normally happens when the model repeats phrases that occurred previously.
    We count the number of the previously occurred 4-gram inside a sentence as a metric.
    Since the length varies across samples, we normalize the count with the total number of 4-grams in the sentence.

    \textbf{Sentence-level Repetition (Sent-REP-4):}
    We define sentence-level repetition as the repetition \textbf{inside the corpus}, which accounts for dull and generic sentences.
    We first collect all the 4-grams of the headlines that appear more than once in the training set.
    Then we count the number of occurrences of those 4-grams in the generated headline.
    This effectively captures the sentence-level repetition, as those repetitive patterns and generic sentences often occur more than once inside the corpus.
    Finally, we report the result normalized with the total number of the 4-grams in that headline.

    \textbf{Unique 4-grams (Unique-4):}
    We follow \citeauthor{DBLP:journals/corr/abs-1802-01345} (2018) to use the number of unique 4-grams to evaluate generation diversity. 

    \section{Results}

    \begin{figure*}[t]
        \centering
        \includegraphics[width=0.9\textwidth]{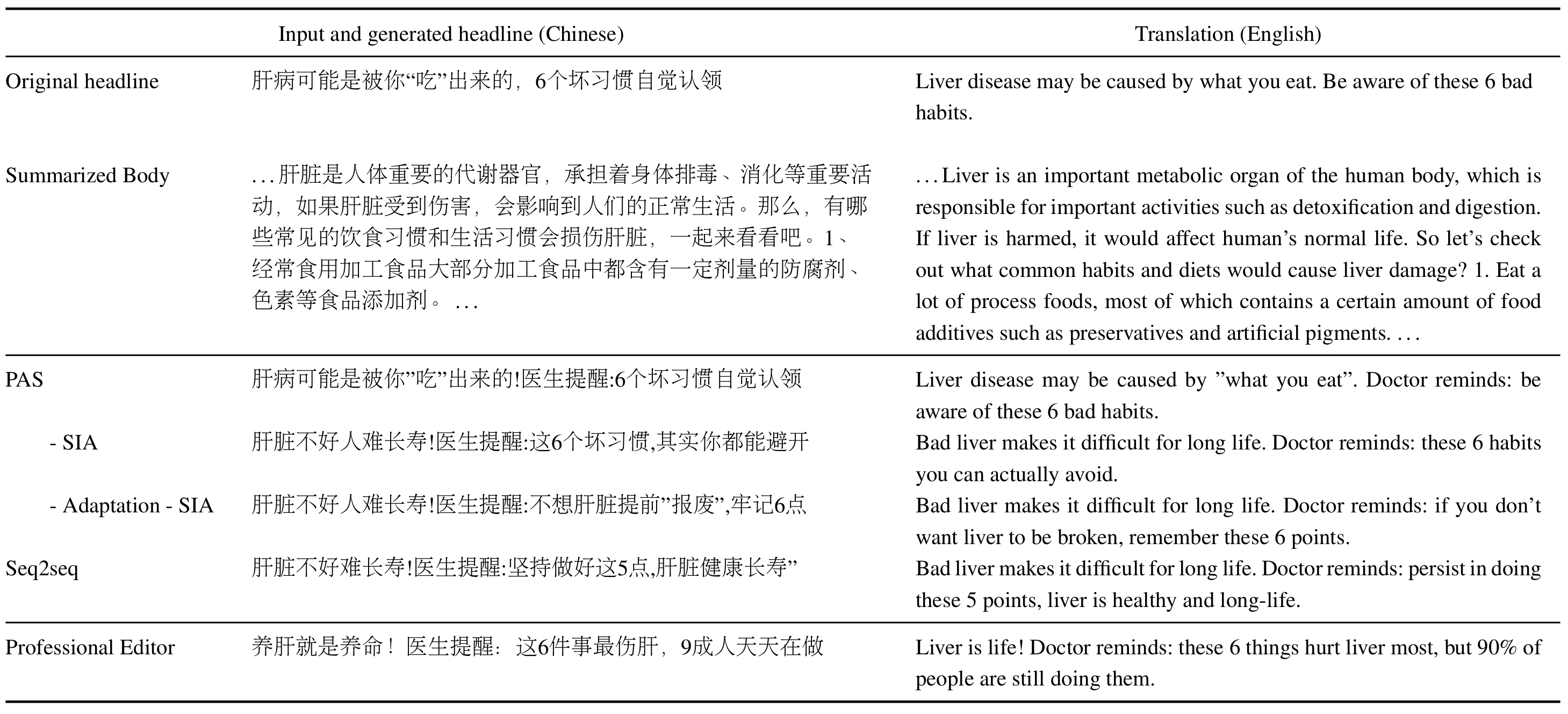}
        \caption{A generated headline example. An original headline and the corresponding news body are used as input. Our method with Adaptation and SIA results in the least repetitive pattern.}
        \label{tab:generation}
    \end{figure*}

    We evaluate the baselines and the proposed model PAS on the PHED dataset.
    Table~\ref{tab:automatic-evaluation} presents the results.
    We calculate the Token-REP-4, Sent-REP-4, and Unique-4 using human written headlines for comparison.
    As we can observe, there is a significant statistical difference between the human and the machine-generated outputs in terms of diversity and repetition.
    Standard sequence-to-sequence with attention (Seq2seq) yields poor results in all metrics, which is reasonable because of the lack of pre-training, suggesting the necessity of using large-scale pre-trained language models to improve the overall quality of generations.
    \hl{Surprisingly, Seq2seq with Adaptation does not work as well as PAS + Adaptation. We suspect that it is probably due to  fact that the two tasks have different inputs (one with the original headline, and the other without). Without pre-training on a more general language model, it seems hard to bridge the headline editing task and the headline generation task for efficient transfer learning.}

    Our full model PAS achieves overall the best results. We further did ablation studies to test each component's contribution on the improvements.
    Note that by default all experiments are conducted on the PHED dataset.
    We can observe that Adaption contributes the most to the improvement on all metrics.
    When removing Adaptation, the performance drops significantly.
    It suggests that adaptation with headline generation task lets the model learn the prior knowledge that helps the headline editing task.


    SIA achieves further improvements in addition to adaptation.
    We set SIA's $\alpha = 0.2$ and $\beta = 40.0$ from the hyper-parameter search results.
    As previously mentioned, MLE cannot deal with the importance imbalance in the dataset.
    In contrast, SIA is aware of such imbalance and adjusts the importance dynamically during the training.
    Compared to MLE, the proposed SIA improves on almost all metrics.
    The results show that SIA not only reduces Token-REP-4 and Sent-REP-4, but also improves on traditional metrics such as BLEU-4 and ROUGE-L, which suggests that SIA is a robust loss function in text generation.

    Among the generated samples, we select a representative example, shown in Table~\ref{tab:generation}, to illustrate the difference among different models.
    From all models, our full model PAS (Pre-training + Adaptation + SIA) has the best result.
    It closely mimics the professionally edited headline style, with a commonly seen pattern ``Doctor reminds:".
    Notice that more of such patterns does not mean better quality, but should depend on the news body.
    When we remove SIA, a repetitive pattern occurred before in the training set ``Bad livers makes it difficult for long life" appears in the generated headline; however, this pattern does not fit the news body here.
    Further removing adaptation, the generated headline produces a logical error
    :``remember these 6 points" conflicting with the true meaning of the news body, which reflects that the model without adaptation relies more on hallucination rather than summarization.
    As for Seq2seq's generated headline, the number predicted is wrong and it also contains repetitions, which results in the worst quality among all models.

    \begin{figure}[h]
        \centering
        \begin{minipage}[b]{0.227\textwidth}
            \includegraphics[width=\textwidth]{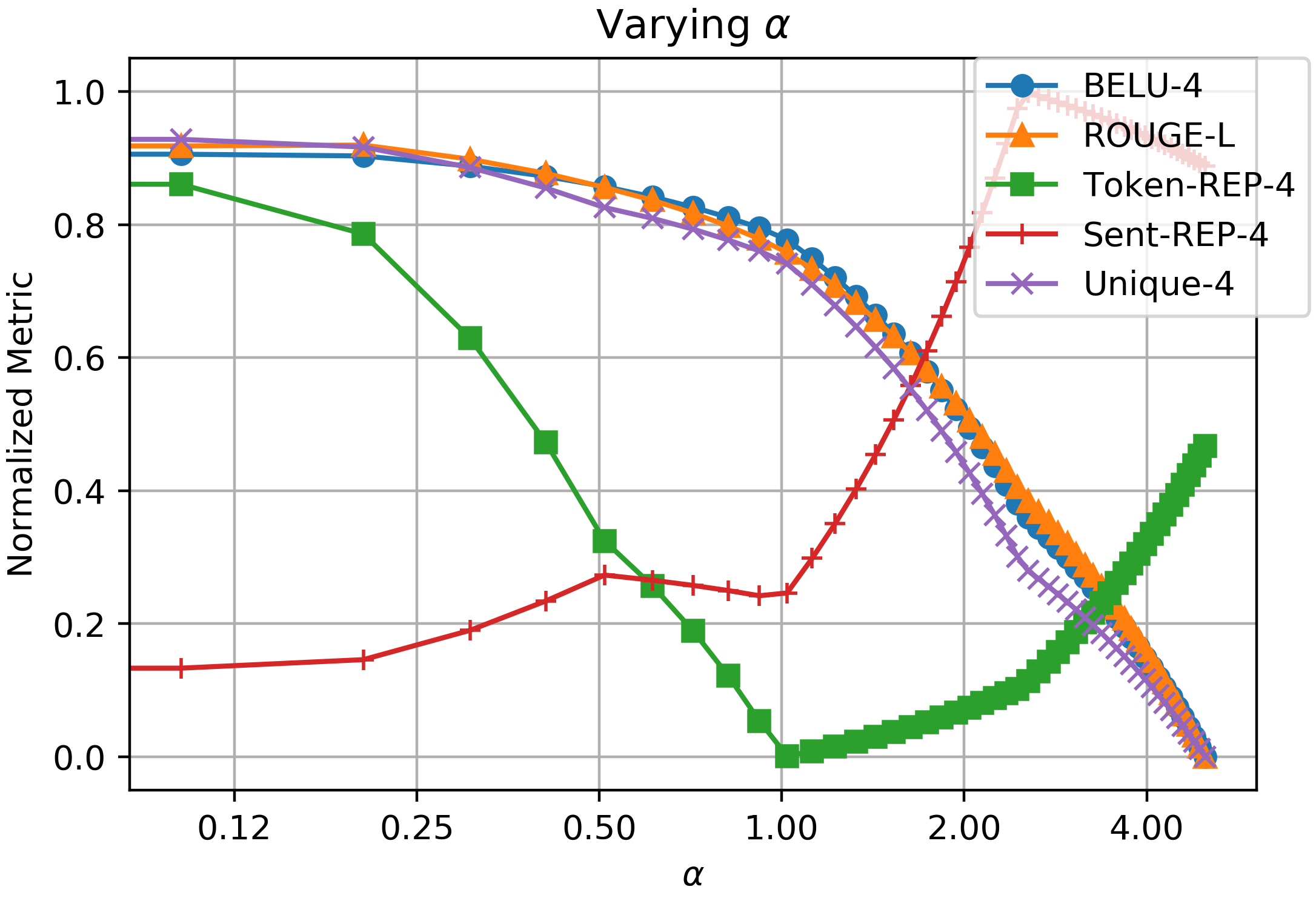}
            \caption*{(a) Varying $\alpha$}
        \end{minipage}
        \hfill
        \begin{minipage}[b]{0.241\textwidth}
            \includegraphics[width=\textwidth]{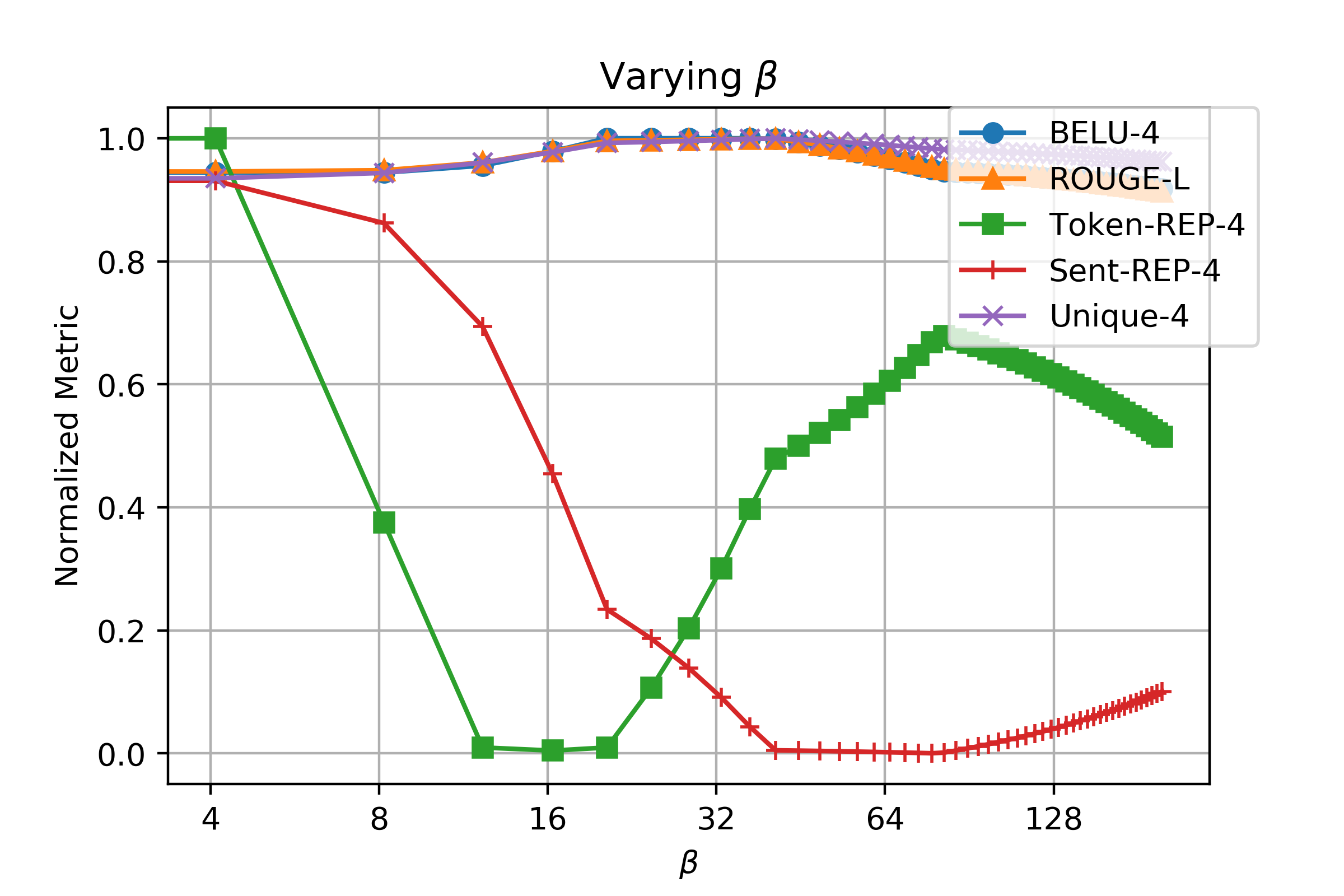}
            \caption*{(b) Varying $\beta$}
        \end{minipage}
        \caption{Effects of varying $\alpha$ and $\beta$. Larger $\alpha$ results in lower Token-REP-4 but hurts BLEU-4. In contrast, larger $\beta$ decreases Sent-REP-4 but improves BLEU-4. Note that all metrics are normalized for better visualization.}
        \label{fig:trend}
    \end{figure}

    \subsection{Reducing Repetition in Token and Sentence Level}

    SIA's two new hyper-parameters $\alpha$ and $\beta$ control the degree of penalty for easily predicted tokens and sentences.
    To explore these parameters' effects on repetitions, we search over different settings and report the performance trend.
    We plot the trends separately for $\alpha$ and $\beta$ in Figure~\ref{fig:trend}.
    We vary $\alpha$ first while setting $\beta=0.0$. $\alpha$ controls how the importance of easily predicted tokens are down-weighted.
    As we enlarge $\alpha$, Token-REP-4 decreases, suggesting fewer 4-grams repetition in a sentence.
    However, large $\alpha$ hurts the performance of all other metrics.
    We suspect that this is because each token probability is dependent on the preceding tokens and, as $\alpha$ becomes larger, it breaks
    the sentence language model distribution, which leads to worse results.
    Also, when $\alpha \ge 1.0$, the Token-REP-4 stops decreasing.
    Therefore, we need be cautious when setting $\alpha$.

    Then we vary $\beta$ to see the effects on the sentence-level repetitions and other metrics.
    The results show that changing $\beta$ is more robust and stable than setting $\alpha$.
    Large $\beta$ penalizes easily predicted sentences' importance.
    As we enlarge $\beta$, Sent-REP-4 decreases as expected.
    It stops decreasing after $\beta \ge 80.0$.
    \hl{Further setting larger beta would not linearly increase the diversity (Token/Sent-REP-4).}
    But even if $\beta$ is set very large, it only leads to slight degeneration on the performance.
    Setting appropriate $\beta$ improves not only Sent-REP-4 but also standard metrics like BLEU-4 and ROUGE-L.
    This further suggests that without detailed searching of $\beta$, we can improve the overall quality of generations using SIA.

    Finally we vary $\alpha$ and $\beta$ jointly to see if they have an interaction effect.
    Based on the experience of separately varying those hyper-parameters, we search a limited range of combinations that $\alpha$ is between $0.1$ and $0.5$ and $\beta$ is between $20.0$ and $40.0$.
    We observe that there is trade-off between Token-REP-4 and other metrics. If we want Token-REP-4 to be too small, it would hurt other metrics.
    Therefore, we set B $\alpha=0.2$ and $\beta=40.0$ as the final hyper-parameters in SIA, because such a parameter setting yields overall best performance.

    \subsection{Human Evaluation}

    \begin{table}[ht]
        \footnotesize
        \begin{center}
            \resizebox{0.46\textwidth}{!}{
                \begin{tabular}{l|c|c|c}
                    \toprule
                    Model 1 vs. Model 2                          & Win    & Lose   & Tie     \\
                    \midrule
                    PAS vs. PAS - SIA                            & 7.56\% & 6.44\% & 86.0 \% \\
                    PAS vs. PAS -Adaptation -SIA                 & 30.0\% & 24.4\% & 45.6\%  \\
                    PAS vs. Original headline                    & 57.4\% & 27.7\% & 14.9\%  \\
                    \midrule
                    PAS vs. PAS - SIA (only different sentences) & 48.5\% & 44.3\% & 7.23\%  \\
                    \bottomrule 
                \end{tabular}
            }
        \end{center}
        \caption{Human evaluation results. ``-" means without using the method.}
        \label{tab:human-evaluation}
    \end{table}

    Automatic evaluation metrics are limited, as they only capture one aspect of the generation quality at a time.
    Therefore, we conduct human evaluation to examine overall generation quality \cite{DBLP:conf/emnlp/NovikovaDCR17}.
    We randomly select 100 generated samples from the test set (1,500 samples in total) and ask five native Chinese speakers to evaluate their quality on different models.
    The results are shown in Table~\ref{tab:human-evaluation}. We observe that the model with SIA is preferred by human judges (17.4\% more favored) over the model without SIA.
    Further removing the adaptation process after removing SIA, the relative preference for PAS increases to 23.0\%.
    Human judges also pointed out that they prefer PAS because the generated headlines have less generic patterns or repetitions.
    Such observation also aligns with the results using automatic evaluation metrics.
    To test our headline editing model's applicability in real-world application, we also compare the full model against the original headlines.
    The results show that our generated headlines are much more favored by humans than the original ones (107\% more).
    We believe PAS has learned the style from professional editors.

    One thing to be aware of is that there is a large percentage of Ties between PAS and PAS without SIA.
    86\% of judges think the two headlines are equal in the quality. This is because if we fix the random seed across experiments, the output distribution from SIA is similar to the one from MLE (since SIA is built on top of MLE). Therefore, the two methods generate a large number of headlines that are exactly the same.
    In order to discover the two methods' difference in depth, we manually filter out headlines that are exactly the same between two models in the test set (1500).
    Among the 234 resulting headlines, we randomly select 100 headlines again and ask judges to compare them.
    The result is in the last row in Table~\ref{tab:human-evaluation}. The percentage of ties between two models drops as expected, and human judges still prefer PAS over PAS without SIA.


    \subsection{Error Analysis}
    All neural models are not perfect (without making any mistakes) and our model is no exception. We perform an error analysis to discover the drawbacks of our model.
    We recruited an expert to rate the appropriateness of the 100 randomly selected generated headlines from the Full model. We find that nearly 53\% of the generated headlines are not considered appropriate given the news body.
    Within this 53\% of samples, 76\% of them are labeled as having comprehension or logical errors.
    Many errors are caused by memes or nicknames, for example Melon is the nickname for Carmelo Anthony by Chinese fans.
    It is difficult for a machine to interpret those memes unless it has a good knowledge base for such nicknamed linkage.
    Fixing logical errors is potentially even harder because current neural networks, especially generative models, perform poorly in reasoning and inference.
    Although those errors may cause the generated headlines to be worse than the original headlines, in the headline recommendation setting, users still have the freedom to choose between their original headline and the machine generated headline, which makes the logic problem less fatal in real-world application. In future work, we aim to design algorithms to detect and tackle these logical errors in generation.


    \section{Conclusion}

    In this paper, we propose a neural headline editing model that aims to generate more attractive headlines with less dull and generic patterns and repetitions.
    To train such model, we construct the professional headline editing dataset (PHED) with the original headlines and the edited headlines collected from professional editors.
    We leverage pre-training with large-scale Chinese corpus and adaptation with a headline generation task before fine-tuning on PHED.
    Experimental results show that the adapted model dramatically decreases the repetitions and improves on other metrics such as BLEU-4 and ROUGE-L.
    Furthermore, we design a self importance-aware (SIA) objective function to be aware of the importance difference of data points during training.
    SIA automatically focus on learning harder tokens and sentences by down-weighting their importance in the loss function.
    The results from SIA show that setting the correct $\alpha$ and $\beta$ can further improve the performance over the model with pre-training and adaptation.
    For future work, we would like to apply SIA to other language generation tasks to analyze the effects.

\bibliography{generation.bib}
\bibliographystyle{aaai}
\end{document}